# A Causal Approach for Business Optimization: Application on an Online Marketplace

Naama Parush[†]
VIANAI Systems
naama@vian.ai

Ohad Levinkron-Fisch[†]
VIANAI Systems
ohad@vian.ai

Hanan Shteingart
VIANAI Systems
hanan@vian.ai

Amir Bar Sela
Worthy
amir.barsela@worthy.com

Amir Zilberman
Worthy
amir.zilberman@worthy.com

Jake Klein
VIANAI Systems
jake@vian.ai

## ABSTRACT

A common sales strategy involves having account executives (AEs) actively reach out and contact potential customers. However, not all contact attempts have a positive effect: some attempts don't change customer decisions, while others might even interfere with the desired outcome. In this work we propose using causal inference to estimate the effect of contacting each potential customer and setting the contact policy accordingly.

We demonstrate this approach on data from Worthy.com, an online jewelry marketplace. We examined the Worthy business process to identify relevant decisions and outcomes, and formalized assumptions on how they were made. Using causal tools, we selected a decision point where improving AE contact activity appeared to be promising. We then generated a personalized policy and recommended reaching out only to customers for whom it would be beneficial. Finally, we validated the results in an A/B test over a 3-month period, resulting in an increase in item delivery rate of the targeted population by 22% (p-value=0.026). This policy is now being used on an ongoing basis.

## 1 INTRODUCTION

Digital business-to-business (B2B) sales processes typically involve a large number of interactions between potential customers and an enterprise. This is also the case for business-to-consumer (B2C) "considered purchases", where customer buying journeys can be complex – for example in the insurance, finance, or telecommunications industries. These interactions can be automated, system driven, or initiated and mediated by sales professionals, such as account executives (AEs) or business development representatives (BDRs)[1,33]. As personnel resources are often limited and relatively expensive, there is a significant business impact associated with improving AE decision making around which potential customers should be contacted, when and how (typically email, telephone, SMS or online chat)[8]. A common industry practice is to employ a lead scoring approach, where each lead is assigned a score depending on certain criteria. In the past, lead scoring involved using predefined metrics and heuristics that are meant to estimate the likelihood of a potential customer converting. More recently, sales organizations have begun to employ predictive lead scoring, where each customer's likelihood to convert is estimated using a machine learning model, and those individuals with a high probability of conversion are then prioritized for sales outreach[18]. This is often done within customer relationship management (CRM) or sales engagement software systems. While the predictive approach focuses sales on customers most likely to convert, it may waste efforts on leads who would convert regardless ("sure things"), or cases where outreach would have an adverse effect ("sleeping dogs")[2,21].

In this work, we address the problem of optimizing AE decisions with a causality-based approach. Collaborating with Worthy.com, an online marketplace for pre-owned jewelry, we identified specific AE decisions to optimize. We then built a policy to generate recommendations for those decisions, and estimated its value, using historical data and tools from the causal inference literature. We approved this policy with stakeholders and evaluated it in a randomized trial (A/B test). Following the success of this evaluation, the policy was rolled out to production and is now used by Worthy AEs.

In the following sections we describe our approach for policy optimization with causal inference (Section 1.1), and the Worthy problem setting (Section 1.2).

### 1.1 Policy Optimization with Causal Inference

In a setting where an agent can perform several actions, a policy $\pi$ is defined as:

$$\pi(x): x \to A$$

Where $x$ is a feature space representing the context at a given decision (also called "effect modifiers"), and $A$ is an action. Our goal is to produce a policy that maximizes an objective. In our case, the AE is represented as an agent, the action is whether or not to contact a lead (an individual that may be interested in selling a piece of pre-owned jewelry), and a simple objective can be defined as the sum of sales over all leads in a time period.

---
[†] Both authors contributed equally to this research.

Leads are assumed to be independent, so that the action on one lead does not affect the sales of other leads ("no interference"[5]). Under this assumption, decisions for each lead can be optimized separately.

For a specific decision for a lead, the optimal policy should recommend the action that maximizes the probability of a sale:

$$\pi_{optimal}(lead) = \underset{A}{\mathrm{argmax}}\, P(sales_A)$$

Where the quantity $P(sales_A)$ represents the probability of a sale if the decision were to take action $A$.

$sales_A$ is a potential outcome: for a given lead, only a single action is chosen and a single outcome $sales_A$ is realized. This is called the "factual" outcome, and the values of $sales_A$ for other actions are called "counterfactual" and can never be observed[1].

In the simple case of a binary action, we can define a (counterfactual) quantity called the "effect" of the action on sales, as the difference in the probability of a sale between doing and not doing $A$. For a given lead, we can estimate the effect based on the features $x$ that describe the lead. This is known as the conditional average treatment effect (CATE):

$$CATE(x) = P(sales_{A=1}|x) - P(sales_{A=0}|x)$$

The policy that then optimizes the objective above is simply to do $A$ if the CATE is positive, and not do $A$ if it's negative.

Note that, in contrast to supervised learning, this setting has no "label", as we never observe $\underset{A}{\max}\, P(sales_A)$ or the effect of $A$ on sales. Nevertheless, the counterfactual "label" can be estimated using data of leads with different values of $A$, assumptions on the data generating process for $A$ and $sales_A$, and tools from the causal inference literature.

If we could conduct a well-designed randomized experiment (A/B test), estimating the CATE would have been relatively straightforward, since we assume contacted and non-contacted leads could be compared without bias (are "exchangeable"[27]). However, collecting enough randomized data to train a policy is challenging, as it requires many decisions to be made randomly, resulting in significant business impact. Instead, we leverage data from existing past decisions.

Recently, there have been significant advances in estimating CATE, in particular from non-randomized data. Developments include theoretical frameworks (e.g., [22, 26]), guidelines on how to approach such problems (e.g., [16, 17, 24, 27]), specialized machine learning techniques for efficient estimation (e.g.,[14, 31, 37]), and maturing open source tools[4,9,32]. In parallel, in the reinforcement learning community new methods have been developed for estimating the value of policies on offline, non-randomized data (e.g.,[13, 19, 35]).

To our knowledge, this work is the first to show the value of a policy built from observational data, verified experimentally, in the domain of sales and marketing. The field of marketing has increasingly embraced modeling effects (or "uplift"[15]) for personalized decision making, though largely limited to data from randomized experiments[12, 21]. Decades of econometric literature has dealt with decision making based on historical data for retail pricing & promotion (e.g.[29]), and more recently for advertising[6,7,10], nonetheless we did not find experimental demonstrations of the value of the proposed policies.

### 1.2 The Worthy.com Use Case

Worthy.com is an online marketplace for pre-owned jewelry, connecting sellers (here referred to as leads) with potential buyers. After the leads register their items online, they are asked to ship their valuables to Worthy offices for grading and evaluation. Later, it is auctioned to potential buyers pending final approval from the lead. As part of the Worthy sales process, AEs reach out to eligible leads (above a minimal value estimate, non-fraudulent etc.) in an effort to increase sales by alleviating some of the friction and walking the lead through the process.

Item delivery is a critical point towards a successful sale and is the main stage at which AE contact decisions have an impact. Subsequent stages involve objective assessments of the potential deal by both buyers and sellers and are less affected by AE actions and capabilities. Therefore, a successful outcome of the AE communication policy is the delivery of the item to the company's offices. Previously, the policy AEs followed was to call potential sellers on the first day an item is registered online. Depending on whether the call went through and its content, the next contact attempt is then scheduled according to a predictive lead score, "soft" guidelines and the assigned AE's experience and intuition. In addition, the lead continues receiving automatic messages during the time between calls.

## 2 METHODS

### 2.1 Problem Framing

Our project focuses on AE contact decisions. This is according to Worthy's suggestion, as this step was perceived to have a significant impact on leads and could benefit from optimization and personalization. When considering how and when to intervene with the AE decisions we first define the scope of our intervention:

1. The new policy will only apply to users assigned to actual AEs and not only to automatic systems.
2. Only the first contact of the day involves a decision based on the contact policy. Subsequent communications are the result of the first one.
3. If the lead initiated the first communication of the day, it is not a result of the contact policy. In the same manner, if leads prescheduled phone calls for the following days,

---

[1] Here we use the potential outcomes notation[28]. Equivalently we could describe the problem using the $do(a)$ operator of the structural causal models framework[23].

they will be contacted on the set timeslot and not otherwise.

4. Since AEs have many tasks on their table when we recommend contacting a lead, we cannot control the exact timing during the day. Therefore, we assume the recommendation will be given at the beginning of the day and carried out later according to the AE's schedule.
5. We limited our work to eligible, non-commercial (private) leads, with a single item, and to popular jewelry types.

## 2.2 Observational Data

We conducted an observational study on data that were collected over three years between April 2018 and May 2021. Following section 2.1, the decision times were defined as the beginning of the AE shift at 9 AM of each morning. The outcome was defined as whether the item was delivered within 3 months from registration. As for the action, if a lead was not contacted for the entire day, the action is defined as "no contact". If the first communication of the day was initiated by the AE, the action is defined as "contact". Otherwise, if the lead initiated communication first on that day, the action was undefined (and excluded from the analysis).

## 2.3 Choosing the Intervention Day

During the lifetime of the AE and lead's relationship there are many contact decisions. According to the decision points we defined above, the number of decisions is equal to the number of days the item is active. In addition, the effect of contacting depends on how long the item is active (e.g., leads on registration day are impacted differently than leads a week later). Therefore, similarly to [36], we can view this decision process as nonstationary, where each day is an episode, and transition matrices change between episodes. We then train a separate model for each day within the process. In addition, not all days are equally important. The optimized policy for specific days might have added value over the existing AE policy, while for other days we may not be able to improve by much. Therefore, we evaluated several decision days and chose to focus on the policy that predicted the highest added value. We excluded the day of item registration ("day 0"), since each item was registered at a different time throughout the day and most items would not exist when the recommendation is given. Subsequent days are denoted as "day 1" (the day following registration), "day 2" etc.

## 2.4 Assumptions on the Underlying Process

We list the features involved in the AE contact decision, and causes of delivery:

1. Lead characteristics – information about the lead (such as the lead's state and age)
2. Item characteristics – information about the registered item (such as the value estimation)
3. Item Worthy stage – internal stage indicating the progress that the lead made since submitting their information on the website (such as "ready to ship")
4. Communication history – all the messages and phone calls prior to the decision. There are two feature types: first, the communication times and quantities and second, communication content. While the former was available for us, the latter was not included in our data but known to the AE at decision time
5. Submission time/date/season – features independent of the lead/item that can affect the decision and outcome
6. AE – the AE's personal experience, preferences, and strategy. Since the AEs were replaced over time and in addition in many cases we could not know the AE identity at the decision time, this feature was not used in our current model

All of the above features affect the decision whether to contact the lead, and together with the contact action also affect the delivery outcome, as illustrated in Figure 1 in Supplementary.

## 2.5 CATE Estimation

According to 2.4, we adjust for all available causes of the action and outcome.
Building a CATE estimator consisted of the following steps:

1. Feature selection: We used feature selection for uplift modeling[38] to select the leading 50 features.
2. Positivity: In order to assure every action has a non-zero chance of being selected, i.e. maintain positivity, we calculated the propensity score (PS) for each item (the probability to be contacted) and excluded items with PS<0.01 or PS>0.99 from the dataset[27]. Less than 1% of the data was excluded.
3. Training the effect estimator: To predict the effect, we trained an ensemble of 30 uplift random forests[30] using the CausalML implementation[9] and averaged over all predictions as the final estimate.
4. Quantifying model performance: As a proxy for the CATE "ground truth", we binned samples by the predictions from our model, and estimated the average effect within each bin. The calibration plot then shows the CATE predicted by the model against these ground truth proxies[11]. Likewise, we show the Qini curve[25] (similar to the ROC curve for predictions). We also calculate the Qini coefficient, where larger values correspond to better discrimination between negative and positive CATEs across all thresholds (similar to ROC AUC).

## 2.6 From model to policy

We define a policy by setting a threshold over the CATE estimates:

$$\pi_{th}(\tau) = \begin{cases} contact, & \tau \geq th \\ no\ contact, & \tau < th \end{cases}$$

Where $\tau$ is the effect estimate and $th$ the threshold. Since we aim to contact only when the CATE is positive, we use $th = 0$.

## 2.7 Off-policy policy evaluation (OPE)

For estimating the value of the new policy, we used the self-normalizing importance sampling method [34]. For each threshold we evaluate the value of the corresponding policy and plot the OPE as a function of the percent of the population contacted. This plot begins with the OPE of the policy of not contacting all items and ends with the OPE of contacting all items.

## 2.8 Experimental validation of the results

After the OPE value indicates that the new policy improves over the existing one, we validate the result with a randomized controlled trial (A/B test). As compared to a randomized trial for training a policy (without leveraging prior data), this validation experiment has a much smaller business risk, since vastly less decisions are required. Moreover, we are not comparing different actions, but rather different policies: the existing policy and the optimized one.

For this experiment we use delivery within 2 weeks as a short-term proxy for the long-term 3 month delivery outcome. This is common practice at Worthy, since roughly 70% of the delivered items were delivered within 2 weeks.

The experiment is carried out on a random subset of the eligible items. For those items, we generate the policy recommendations. 50% of the items then receive these recommendations (treatment group) and 50% continue with the existing policy (control group). Recommendations were integrated into the existing contact management system, and AEs were not aware whether the new was involved.

## 3 RESULTS

### 3.1 Observational analysis

We analyzed data that consisted of over 200,000 items. Out of the 3-year period, the last 3 months of data were held out for validation. As described in section 2, we generated CATE estimates and corresponding policies for the different days and estimated the value of optimizing decisions at each day. Figure 1a,b show the CATE calibration and normalized Qini curve for day 1. These plots show good separation between positive and negative CATE, and a normalized Qini coefficient of 0.71. For comparison, we trained a predictive model for the same outcome, resulting in a ROC AUC of 0.849, and the corresponding Qini coefficient was not greater than random. Normalized Qini coefficients for policies of subsequent days were lower than day 1 but larger than random (e.g. day 2=0.31, day 3=0.3). Figure 2 shows the OPE plot of day 1 policy. This policy was found to provide high added value over the existing policy (29.06% vs. 16.78%). The day 1 OPE for the 2-week outcome showed similar results (23.82% vs. 12.25%) (see Figure 2 in Supplementary). The OPE for the predictive model showed no benefit. The policies of subsequent days did not predict substantial additional value.

Based on the results above, we suggested using the policy for day 1 decisions, contacting leads with a CATE estimate > 0. This policy recommends contacting 66% of the leads, while under the existing policy only 22% of leads were contacted.

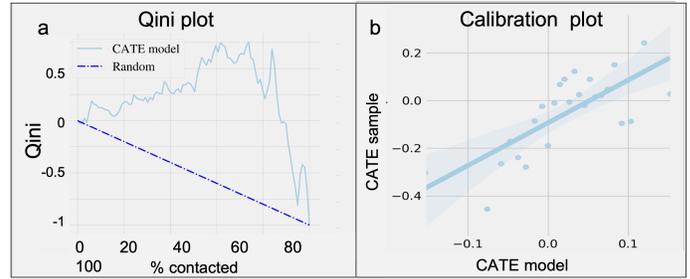

Figure 1: a. Normalized Qini curve (coefficient=0.71); b. CATE calibration plot

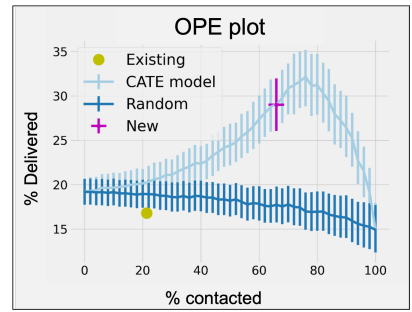

Figure 2: OPE plot for items ordered by i) CATE model estimates ii) random order. The value of the existing policy and the new policy are marked in yellow and purple. Error bars indicate 95%CI.

For explainability reasons, we generated a simple decision tree that imitates the policy (see Figure 3 in Supplementary). This is important for sharing the policy with the domain experts for purposes of receiving feedback and for integrating the policy into AE activities.

### 3.2 Experiment

The experiment was carried out over 3 months, from August 24 to December 1, 2021. A short time before the experiment started Worthy added an additional stage to the registration process and only users that indicated they were interested in selling their item in the near future were assigned to AEs. In addition, pre-scheduling was made more accessible. As a result, the experiment group was somewhat different than the training data.

As in the training process, we excluded from the experiment items where the lead initiated or prescheduled the first contact. In addition, we excluded items where the assigned AEs were absent on the recommendation day (which led to excluding all weekends). In total there were 1781 items in the control group and 1722 in the treatment group (p-value for sample ratio mismatch[20] 0.32, >0.01 so no SRM detected) (Table 1 in Supplementary).

When analyzing the results, we must keep in mind that the AEs in the treatment group could choose to ignore our recommendation and we couldn't entirely control their activity. In practice they followed our recommendation for 54% of the items. The control group were unaware of our recommendations, however for 29% of the items, their actions were in line with our policy. The treatment group delivery rate (10.52% [95% CI 9.11%, 12.02%]) was 1.91% higher than the control group (8.61% [95% CI 7.3%, 9.99%]), a relative increase of 22% (one sided t-test, p-value=0.026).

The contact rate in the treatment group was 51%, compared to 36% in the control group. To get an indication for whether the improvement was caused by merely increasing the contact rate, we repeated the analysis separately for items with "contact" and "no contact" recommendations. The results in Table 1 in Supplementary show that the improvement is not confined to "contact" recommendations.

## 4 CONCLUSIONS

This work demonstrates the value of a causality-based framework for optimizing business decisions. We identified a decision to optimize and explored how past decisions were made. We then used historical data to estimate the effect of contacting a given lead and generated a corresponding policy. As this policy showed substantial added value over the existing one we set out to test it in a randomized experiment. The success of the experiment, led to its adoption as the standard practice today.

Our analysis shows that contacting more is not necessarily better. While our policy does indeed recommend a higher contact rate, contacting all leads is predicted to impair the overall delivery rate (Figure 1). This is also evident in the "non-contact" recommendations group in the experiment, where less contacts resulted in more deliveries.

Another aspect of this study is the integration of human decisions with artificial intelligence (AI). While the AEs' policy is not statistically optimized, the recommended policy does not have the full information at decision time (such as the prior calls content). The AE chooses whether to accept the recommendation and in practice integrates both policies. This integration can be synergetic or sub-optimal. At this point we are not able to determine if complying blindly to the new policy would have improved the results or if the AEs decided not to comply when the policy was incorrect, and only conclude that the integrated policy performed better than the existing one.

We point out the difference between policy value and CATE accuracy. Even though we could estimate CATE values for days 2 and onwards, the resulting policies did not exhibit substantial benefit over existing practice. By excluding these days, we reduced noise and were able to achieve conclusive results.

Our results are limited in several aspects: first, we optimized for item deliveries and not directly for sales or profit. While this was researched at Worthy and is standard practice, evaluating whether this served as a good short-term proxy[36] was outside the scope of this work. Additionally, our analysis of the business process (Section 2.4) concluded there may still be significant hidden confounding, resulting in biased policy and OPE estimates. While this does not detract from the results of the experiment, it does leave room for improvement. Finally, as customer and firm behavior changes considerably over the course of months or weeks, the benefits we show may not last, and periodic improvements may be required.

Future work may include incorporating additional information about AEs and communication history, to reduce hidden confounding. Joint optimization of the sequence of decisions could also be explored, integrating methods from the reinforcement learning literature. In addition, mechanisms for dealing with ongoing changes in the business environment should be developed.

Finally, while here we focused on optimization, another important benefit could be automating AE decisions, reducing the need to manually plan contact actions and shortening training for new AEs.

## 5 ACKNOWLEDGMENTS

We thank the Vianai and Worthy teams and Becca Feldman for insightful comments, and Ronny Kohavi for consulting with us on experimentation.

# SUPPLEMENTARY

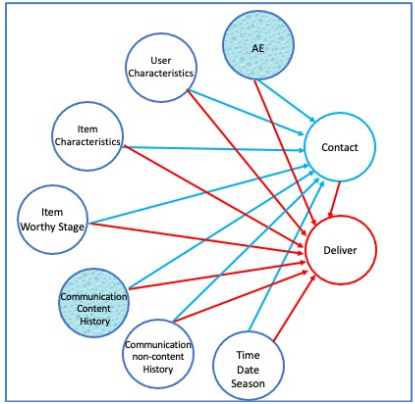

**Figure 1: Causal graph of the contact decision and delivery outcome. Filled circles indicate missing information.**

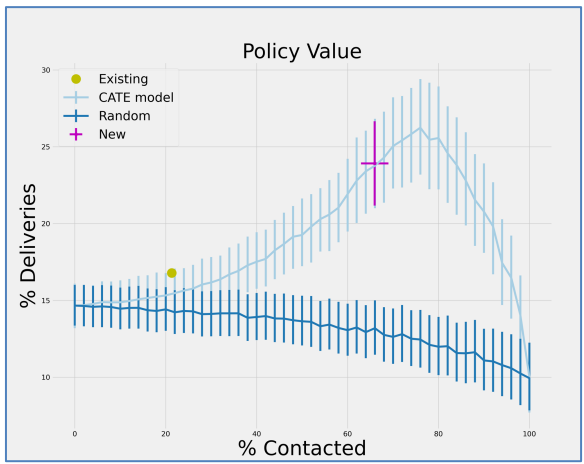

**Figure 2: 2-week proxy outcome OPE plot for items ordered by i) CATE model estimates ii) random order. The value of the existing policy and the new policy are marked in yellow and purple. Error bars indicate 95%CI.**

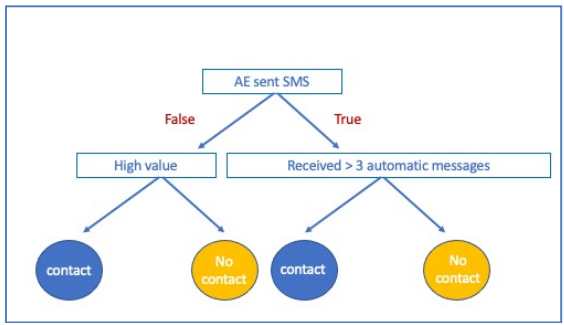

**Figure 3: A simplified decision tree describing the new policy**

**Table 1: Experiment results for the control and treatment groups, and for the group of items that received "contact" recommendation and the group that received "no contact" recommendation**

|  | Control | | | Treatment | | | Effect (difference in delivery rate) |
|---|---|---|---|---|---|---|---|
|  | # Items | Delivery Rate | Compliance Rate | # Items | Delivery Rate | Compliance Rate |  |
| All items | 1781 | 8.6% | 29% | 1722 | 10.5% | 54% | 1.9% |
| "Contact" recommendations | 1248 | 7.4% | 25% | 1236 | 9.2% | 53% | 1.8% |
| "No contact" recommendations | 533 | 11.4% | 38% | 486 | 13.8% | 55% | 2.4% |